\documentclass[11pt,a4paper]{article}
\usepackage[hyperref]{naaclhlt2018}
\usepackage{times}
\usepackage{latexsym}
\usepackage{url}
\usepackage{tikz}
\usetikzlibrary{quotes,angles}
\usepackage{danudefs}
\usepackage{graphicx}
\usepackage{tabularx}
\usepackage{pgf}

\usepackage{mathtools}
\usepackage{array}
\usepackage{amsmath, amssymb, amsfonts}

\usepackage{array, multirow, booktabs}

\aclfinalcopy 
\def\aclpaperid{757} 

\title{Frustratingly Easy Meta-Embedding -- Computing Meta-Embeddings by Averaging Source Word Embeddings}

\author{Joshua N Coates \\
  Department of Computer Science \\
  University of Liverpool \\
  {\tt j.n.coates@liverpool.ac.uk} \\\And
  Danushka Bollegala \\
  Department of Computer Science \\
  University of Liverpool \\
  {\tt danushka@liverpool.ac.uk} \\}
  
\newcommand{\setone}{{\cS_{1}}}
\newcommand{\settwo}{{\cS_{2}}}

\newcommand{\wordone}{\vec{u}}
\newcommand{\wordtwo}{\vec{v}}

\newcounter{magicrownumbers}

\date{}

\begin{document}
\maketitle
\begin{abstract}
Creating accurate meta-embeddings from pre-trained source embeddings has received attention lately.
Methods based on global and locally-linear transformation and concatenation have shown to produce accurate meta-embeddings.
In this paper, we show that the arithmetic mean of two distinct word embedding sets yields a
performant meta-embedding that is comparable or better than more complex meta-embedding learning methods.
The result seems counter-intuitive given that vector spaces in different source embeddings are not comparable and cannot be simply averaged. We give insight into why averaging can still produce accurate meta-embedding despite the incomparability of the source vector spaces.
\end{abstract}

\section{Introduction}

Distributed vector representations of words, henceforth referred to as word embeddings, have been shown to exhibit strong performance on
a variety of NLP tasks~\cite{turian2010word, zou2013bilingual}. Methods for producing word embedding sets exploit the distributional
hypothesis to infer semantic similarity between words within large bodies of text, in the process they have been found to additionally capture
more complex linguistic regularities, such as analogical relationships~\cite{mikolov2013linguistic}. A variety of methods now exist for the
production of word embeddings~\cite{collobert2008unified, mnih2009scalable, huang2012improving, pennington2014glove, mikolov2013efficient}. Comparative work has illustrated a variation in performance between methods across evaluative
tasks~\cite{chen2013expressive, yin2015learning}.

Methods of ``meta-embedding'', as first proposed by~\citet{yin2015learning}, aim to conduct a complementary combination of  information
from an ensemble of distinct word embedding sets, each trained using different methods, and resources, to yield an embedding set with
improved overall quality.

Several such methods have been proposed. 1\texttt{TO}N~\cite{yin2015learning}, takes an ensemble of $K$ pre-trained word embedding sets, and employs a linear neural network to learn a set of meta-embeddings along with $K$ global projection matrices, such that through projection, for every word in the meta-embedding set, we can recover its corresponding vector within each source word embedding set. 1\texttt{TO}N+~\cite{yin2015learning}, extends this method by predicting embeddings for words not present within the intersection of the source word embedding sets. An unsupervised locally linear meta-embedding approach has since been taken~\cite{bollegala2017think}, for each source embedding set, for each word; a representation as a linear combination of its nearest neighbours is learnt. The local reconstructions within each source embedding set are then projected to a common meta-embedding space.

The simplest approach considered to date, has been to concatenate the word embeddings across the source sets~\cite{yin2015learning}.
Despite its simplicity, concatenation has been used to provide a good baseline of performance for meta-embedding.

A method which has not yet been proposed is to conduct a direct averaging of embeddings. The validity of this approach may perhaps not seem obvious, owing to the fact that no correspondence exists between the dimensions of separately trained word embedding sets.
In this paper we first provide some analysis and justification that, despite this dimensional disparity, averaging can provide an approximation of the performance of concatenation without increasing the dimension of the embeddings. We give empirical results demonstrating the quality of average meta-embeddings. We make a point of comparison to concatenation since it is the most comparable in terms of simplicity, whilst also providing a good baseline of performance on evaluative tasks. Our aim is to highlight the validity of averaging across distinct word embedding sets, such that it may be considered as a tool in future meta-embedding endeavours.

\section{Analysis}

To evaluate semantic similarity between word embeddings we consider the Euclidean distance measure.
For $\ell_{2}$ normalised word embeddings, Euclidean distance is a monotonically decreasing function of the cosine similarity, which is a popular choice in NLP tasks that use word embeddings such as semantic similarity prediction and analogy detection~\cite{Levy:TACL:2015,Levy:CoNLL:2014}. 
We defer the analysis of other types of distance measures to future work.
By evaluating the relationship between the Euclidean distances of pairs of words in the source embedding sets and their corresponding Euclidean distances in the meta-embedding space we can obtain a view as to how the meta-embedding procedure is combining semantic information. We begin by examining concatenation through this lens, before moving on to averaging.

\subsection{Concatenation}
We can express concatenation by first zero-padding our source embeddings, before combining them through addition.

Without loss of generality, we consider both concatenation and averaging over only two source word embedding sets for ease of exposition.
Let $\setone$ and $\settwo$ be unique embedding sets of real-valued continuous embeddings. We make no assumption that $\setone$ and $\settwo$ were trained using the same method or resources. Consider two semantically similar words $\wordone$ and $\wordtwo$ such that $\wordone, \wordtwo \in \setone$ $\cap$ $\settwo$. Let $\wordone_\setone$ and $\wordtwo_\setone$, and $\wordone_\settwo$ and $\wordtwo_\settwo$ denote the specific word embeddings of $\wordone$ and $\wordtwo$ within the embeddings $\setone$, and $\settwo$ respectively. 

Let the dimensions of embeddings $\setone$, and $\settwo$ be denoted $d_\setone$, and $d_\settwo$ respectively. We zero-pad embeddings from $\setone$ by front-loading $d_\settwo$ zero entries to each word embedding vector. In contrast, we zero-pad embeddings from $\settwo$ by adding $d_\setone$ zero entries to the end of each embedding vector. The resulting embeddings from $\setone$ and $\settwo$ now share a common dimension of $d_\setone + d_\settwo$. Denote the resulting embeddings of any word $\wordone \in \setone \cap \settwo$, as $\wordone_{\setone}^{zero}$ and $\wordone_{\settwo}^{zero}$ respectively.
Now, combining our source embeddings through addition we obtain equivalency to concatenation.
\begin{align} \wordone_{\setone}^{zero} + \wordone_{\settwo}^{zero} = \begin{bmatrix}
\wordone_{\settwo_{(1)}} \\
\wordone_{\settwo_{(2)}}  \\
\vdots \\
\wordone_{\settwo_{(d_\settwo)}} \\
\wordone_{\setone_{(1)}} \\
\wordone_{\setone_{(2)}}  \\
\vdots \\
\wordone_{\setone_{(d_\setone)}} 
\end{bmatrix}
= \begin{bmatrix}
\wordone_\settwo \\
\wordone_\setone
\end{bmatrix}
\end{align}
Note that the zero-padded vectors are orthogonal.

Let the Euclidean distance between these words in each embedding be denoted by $E_\setone$ and $E_\settwo$. Note that for any vector $\wordone \in \mathbb{R}^n$ the addition of zero-valued dimensions does not affect the value of its $\ell_{2}$-norm. So we have
\begin{align}
& E_\setone = \norm{\wordone_\setone - \wordtwo_\setone}_{2} = \norm{\wordone_\setone^{zero} - \wordtwo_\setone^{zero}}_{2} \\
& E_\settwo = \norm{\wordone_\settwo - \wordtwo_\settwo}_{2} = \norm{\wordone_\settwo^{zero} - \wordtwo_\settwo^{zero}}_{2}
\end{align}
Consider the Euclidean distance between $\wordone$ and $\wordtwo$ after concatenation.
\begin{align*}
& E_{CONC} \\
& = \norm{ \begin{bmatrix}\wordone_\settwo \\ \wordone_\setone\end{bmatrix} - \begin{bmatrix}\wordtwo_\settwo \\ \wordtwo_\setone\end{bmatrix} }_{2} \\ 
& = \norm{(\wordone_\setone^{zero} + \wordone_\settwo^{zero}) - (\wordtwo_\setone^{zero} + \wordtwo_\settwo^{zero})}_{2} \\
& = \norm{(\wordone_\setone^{zero} - \wordtwo_\setone^{zero}) - (\wordtwo_\settwo^{zero} - \wordone_\settwo^{zero})}_{2} \\
& = \sqrt{ (E_{\cS_{1}})^{2} + (E_{\cS_{2}})^{2} - 2E_{\cS_{1}}E_{\cS_{2}}cos(\theta) } \\
& = \sqrt{ (E_{\cS_{1}})^{2} + (E_{\cS_{2}})^{2} - 2E_{\cS_{1}}E_{\cS_{2}} (0) } \\
& = \sqrt{ (E_{\cS_{1}})^{2} + (E_{\cS_{2}})^{2} }
\end{align*}
For any two words belonging to the resultant embedding obtained by concatenation, the distance between these words in the resultant space is the root of the sum of squares of Euclidean distances between these words in $\setone$ and $\settwo$.

%
%
%

\subsection{Average word embeddings}
\label{sect:avg-sem}

\par Here we now make the assumption that $\setone$ and $\settwo$ have common dimension $d$.\footnote{Without loss of generality, source embeddings with different dimensionality can be appropriately padded to have the same dimensionality.}
\par Despite there being no obvious correspondence between dimensions of $\setone$ and $\settwo$ we can show that the average embedding set retains semantic information through preservation of the relative distances between words.

Consider the positioning of words $\wordone$, and $\wordtwo$ after performing a word-wise average between the source embedding sets. The Euclidean distance between $\wordone$ and $\wordtwo$ in the resultant meta-embedding is given by
\begin{align*}
& E_{AVG} \\
& = \norm{\frac{(\wordone_\setone + \wordone_\settwo)}{2} - \frac{(\wordtwo_\setone + \wordtwo_\settwo)}{2}}_{2} \\
& = \frac{1}{2}\norm{(\wordone_\setone - \wordtwo_\setone) - (\wordtwo_\settwo - \wordone_\settwo)}_{2} \\ 
& \propto \sqrt{ (E_{\cS_{1}})^{2} + (E_{\cS_{2}})^{2} - 2E_{\cS_{1}}E_{\cS_{2}}\cos(\theta) }
\end{align*}

Now in this case, unlike concatenation, we have not designed our source embedding sets such that they are orthogonal to each other, and so it seems we are left with a term dependant on the angle between $(\wordone_\setone - \wordtwo_\setone)$ and $(\wordtwo_\settwo - \wordone_\settwo)$.
However, \newcite{cai2013distributions} showed that, if $\cX$ is a set of random points $\in \mathbb{R}^{n}$ with cardinality $|\cX|$, then the limiting distribution of angles, as $|\cX| \rightarrow \infty$, between pairs of elements from $\cX$, is Gaussian with mean $\pi/2$. In addition, \newcite{cai2013distributions} showed that the variance of this distribution shrinks as the dimensionality increases.

Word embedding sets typically contain in the order of ten thousand or more points, and are typically of relatively high dimension. Moreover, assuming the difference vector between any two words in an embedding set is sufficiently random, we may approximate the limiting Gaussian distribution described by~\citet{cai2013distributions}. In such a case the expectation would then be that the vectors $(\wordone_\setone - \wordtwo_\setone)$ and $(\wordtwo_\settwo - \wordone_\settwo)$ are orthogonal, leading to the following result. 
\begin{align}
\label{eq:avg}
\mathbb{E} [ E_{AVG} ] = \frac{1}{2} \sqrt{ (E_{\cS_{1}})^{2} + (E_{\cS_{2}})^{2} } \propto E_{CONC}
\end{align}
To summarise, if word embeddings can be shown to be approximately orthogonal, then averaging will approximate the same information as concatenation, without increasing the dimensionality of the embeddings.

\section{Experiments}

\par We first empirically test our theory that word embeddings are sufficiently random and high dimensional, such that they are approximately all orthogonal to each other. We then present an empirical evaluation of the performance of the meta-embeddings produced through averaging, and compare against concatenation.

\subsection{Datasets}
\label{sect:datasets}
We use the following pre-trained embedding sets that have been used in prior work on meta-embedding learning~~\cite{yin2015learning,bollegala2017think} for experimentation.
\begin{itemize}
\item {\bf GloVe}~\cite{pennington2014glove}. 1,917,494 word embeddings of dimension 300.
\item {\bf CBOW}~\cite{mikolov2013distributed}. Phrase embeddings discarded, leaving 929,922 word embeddings of dimension 300.
\item {\bf HLBL}~\cite{turian2010word}. 246,122 hierarchical log-bilinear~\cite{mnih2009scalable} word embeddings of dimension 100.
\end{itemize}
Note that the purpose of this experiment is not to compare against previously proposed meta-embedding learning methods, but to empirically verify averaging as a meta-embedding method and validate the assumptions behind the theoretical analysis.
By using three pre-trained word embeddings with different dimensionalities and empirical accuracies, we can evaluate the averaging-based meta-embeddings in a robust manner.

We pad HLBL embeddings to the rear with 200 zero-entries to bring their dimension up to 300.
For GloVe, we $\ell_{2}$ normalise each dimension of the embedding across the vocabulary, as recommended by the authors. Every individual word embedding from each embedding set is then $\ell_{2}$-normalised. The proposed averaging operation, as well as concatenation, operate only on the intersection of these embeddings. The intersectional vocabularies GloVe $\cap$ CBOW, GloVe $\cap$ HLBL, and CBOW $\cap$ HLBL contain 154,076; 90,254; and 140,479 word embeddings respectively.

\subsection{Empirical distribution analysis}

We conduct an empirical analysis of the distribution of the angle $\sphericalangle [(\wordone_\setone - \wordtwo_\setone)$, $(\wordtwo_\settwo - \wordone_\settwo)]$ for each pair of datasets. Table~\ref{tbl:angle-table} shows the mean and variance of these distributions, obtained from samples of 200,000 random pairs of words from each intersectional vocabulary. We find that the angles are approximately normally distributed around $\pi/2$.
\begin{table}[h!]
\small
\centering
\begin{tabular}{l | l l}
Embeddings & $\mu$ & $\sigma^2$ \\ 
\midrule
GloVe \& CBOW & 1.5609 & 0.0121 \\
GloVe \& HLBL & 1.5709 & 0.0129 \\
CBOW \& HLBL & 1.5740 & 0.0126 \\
\midrule
\end{tabular}
\caption{Observed distribution parameters.}
\label{tbl:angle-table}
\end{table}

\begin{figure}
\centering
\includegraphics[width=1.0\columnwidth]{{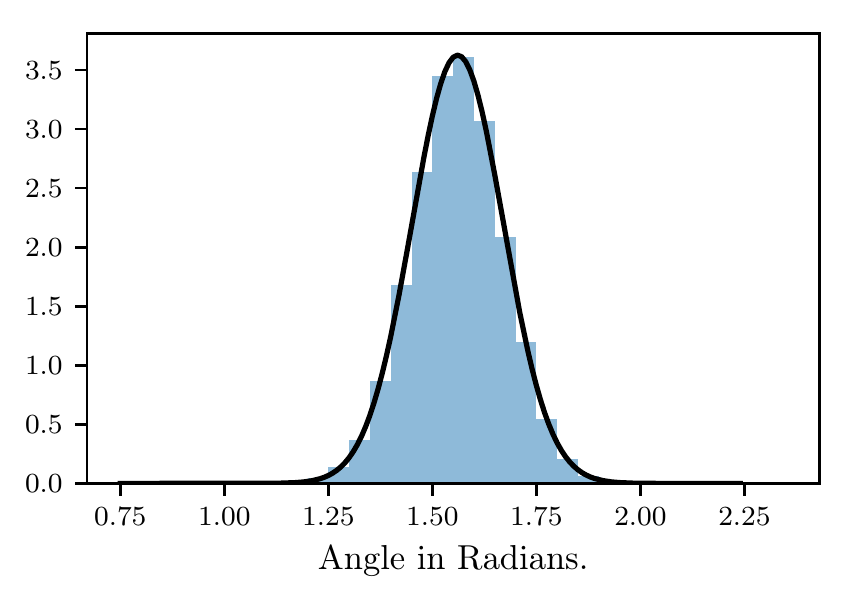}}
\caption{Distribution of angles between embeddings within GloVe $\cap$ CBOW. }
\label{fig:angledistributions}
\vspace{-5mm}
\end{figure}

Figure~\ref{fig:angledistributions} shows a normalised histogram of the results for GloVe $\cap$ CBOW, along with a normal distribution characterised by the sample mean and variance. GloVe $\cap$ HLBL, and CBOW $\cap$ HLBL plots are not shown due to space limitations, but are similarly normally distributed. This result shows that the pre-trained word embeddings approximately satisfy the predictions made by \newcite{cai2013distributions}, thereby empirically justifying the assumption made in the derivation of \eqref{eq:avg}.

\begin{table}[t]
\small
\centering
\begin{tabular}{ p{25mm} p{4mm}  p{4mm} p{4mm} p{4mm} p{4mm} p{4mm}  p{4mm}}
 Embeddings & RG & MC & WS & RW & SL & GL \\ 
\midrule
\textbf{sources} & & & & & & \\
HLBL  100 & 35.3 & 49.3 & 35.7 & 19.1 & 22.1 & 15.0 \\
CBOW 300 & 76.0 & 82.2 & 69.8 & 53.4 & 44.2 & 67.1 \\
GloVe 300 & 82.9 & 87.0 & 75.4 & 48.7 & 45.3 & 68.7 \\
\midrule
\textbf{AVG} & & & & & & \\
CBOW+HLBL 300 & 69.2 & 81.0 & 60.1 & 48.7 & 37.3 & 49.4 \\
GloVe+CBOW 300 & 82.2 & 87.0 & 74.5 & 52.9 & \textbf{46.5} & 73.8 \\
GloVe+HLBL 300 & 73.7 & 74.1 & 64.2 & 44.6 & 38.8 & 49.5 \\
\midrule
\textbf{CONC} & & & & & & \\
CBOW+HLBL 400 & 68.7 & 80.2 & 62.9 & 49.1 & 39.6 & 53.2 \\
GloVe+CBOW 600 & \textbf{83.0} & \textbf{88.8} & \textbf{76.4} & \textbf{54.8} & 46.3 & \textbf{75.5} \\
GloVe+HLBL 400 & 73.7 & 80.1 & 65.5 & 46.4 & 40.0 & 53.8 \\
\end{tabular}
\caption{Results on word similarity, and analogical tasks. Best performances bolded per task. Dimensionality of the meta embedding is shown next to the source embedding names.}
\label{tbl:results-table}
\end{table}

\subsection{Evaluation Tasks}
\label{sext:evaluation}

\subsubsection{Semantic Similarity }
We measure the similarity between words by calculating the cosine similarity between their embeddings; we then calculate Spearman correlation against human similarity scores. The following datasets are used: \textbf{RG} \cite{rubenstein1965contextual}, \textbf{MC} \cite{miller1991contextual}, \textbf{WS} \cite{finkelstein2001placing}, \textbf{RW} \cite{luong2013better}, and \textbf{SL} \cite{Hill2015}.

\subsubsection{Word Analogy}
Using the Google dataset \textbf{GL} \cite{mikolov2013distributed} (19544 analogy questions), we solve questions of the form \textit{a is to b as c is to what?}, using the CosAdd method \cite{mikolov2013linguistic} shown in \eqref{eqn:cosadd}. Specifically, we determine a fourth word \textit{d} such that the similarity between $(b - a + c)$ and $d$ is maximised.
\begin{equation}
\label{eqn:cosadd}
\mathrm{CosAdd}(a:b,c:d) = \cos(b - a + c, d)
\end{equation}

\subsection{Discussion of results}
\label{sect:discussion}

Table~\ref{tbl:results-table} shows task performance for each source embedding set, and for both methods on every pair of datasets. In our experiments concatenation obtains better overall performance. However, averaging offers improvements over the source embedding sets for semantic similarity task \textbf{SL} and word analogy task \textbf{GL}, on the combination of CBOW and GloVe. HLBL has a negative effect on CBOW and GloVe, but the performance of averaging is close to that of concatenation.
An advantage of averaging when compared against concatenation, is that the dimensionality of the produced meta-embedding is not increased beyond the maximum dimension present within the source embeddings, resulting in a meta-embedding which is easier to process and store.

\section{Conclusion}
We have presented an argument for averaging as a valid meta-embedding technique, and found experimental performance to be close to, or in some cases better than that of concatenation, with the additional benefit of reduced dimensionality. We propose that when conducting meta-embedding, both concatenation and averaging should be considered as methods of combining embedding spaces, and their individual advantages considered.

\bibliography{bib-jncoates}
\bibliographystyle{acl_natbib}

\end{document}